\documentclass[conference]{IEEEtran}
\IEEEoverridecommandlockouts
\usepackage{cite}
\usepackage{amsmath,amssymb,amsfonts}
\usepackage{algorithmic}
\usepackage{graphicx}
\usepackage{textcomp}
\usepackage{hyperref}
\usepackage{verbatim}
\usepackage{mathtools}
\DeclarePairedDelimiter\ceil{\lceil}{\rceil}
\usepackage{bbm}
\usepackage[dvipsnames]{xcolor}

\def\BibTeX{{\rm B\kern-.05em{\sc i\kern-.025em b}\kern-.08em
    T\kern-.1667em\lower.7ex\hbox{E}\kern-.125emX}}

\newcommand{\floatfp}{{\sc float32}}

\begin{document}

\title{INT-FP-QSim: Mixed Precision and Formats For Large Language Models and Vision Transformers\\
}

\author{\IEEEauthorblockN{Lakshmi Nair*\thanks{*Email: lakshmi@lightmatter.co},
Mikhail Bernadskiy, Arulselvan Madhavan, Craig Chan, Ayon Basumallik, Darius Bunandar}
\IEEEauthorblockA{Lightmatter Inc.,
100 Summer Street,\\
Boston MA 02110
}}


\maketitle

\begin{abstract}
The recent rise of large language models (LLMs) has resulted in increased efforts towards running LLMs at reduced precision. Running LLMs at lower precision supports resource constraints and furthers their democratization, enabling users to run billion-parameter LLMs on their personal devices. To supplement this ongoing effort, we propose INT-FP-QSim: an open-source simulator that enables flexible evaluation of LLMs and vision transformers at various numerical precisions and formats. INT-FP-QSim leverages existing open-source repositories such as TensorRT, QPytorch and AIMET for a combined simulator that supports various floating point and integer formats. With the help of our simulator, we survey the impact of different numerical formats on the performance of LLMs and vision transformers at 4-bit weights and 4-bit or 8-bit activations. We also compare recently proposed methods like Adaptive Block Floating Point, SmoothQuant, GPTQ and RPTQ on the model performances. We hope INT-FP-QSim will enable researchers to flexibly simulate models at various precisions to support further research in quantization of LLMs and vision transformers.
\end{abstract}


\section{Introduction}
\begin{figure}[t]
\centering
\includegraphics[width=0.48\textwidth]{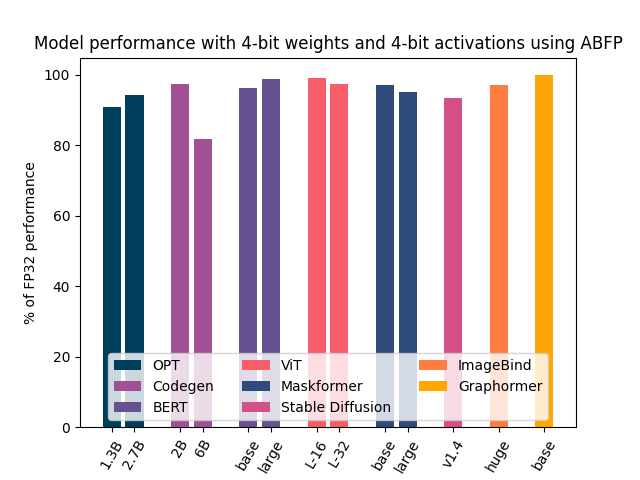}
\caption{Performances of a range of models, relative to FP32, with 4-bit integer weights and activations using Adaptive Block Floating Point (ABFP).}
\label{fig:model_summary}
\end{figure}

The recent rise in popularity of large language models (LLMs) has prompted significant ongoing research into running LLMs at reduced precision, to support resource constraints and democratize their access. Prior work has looked at running the weights and activations of LLMs in 8-bit precision \cite{xiao2022smoothquant,dettmers2022llm}. The most recent techniques focus on enabling 4-bit integer quantization of weights while retaining FP16 activations \cite{frantar2022gptq}, and 4-bit to 3-bit quantization of both weights and activations \cite{yuan2023rptq}. Beyond the recent advancements that have focused specifically on LLMs, numerous techniques have been developed for efficiently running convolutional models and smaller scale language models such as BERT at low precision \cite{dai2021vs,basumallik2022adaptive,wu2020integer}. However, the latter techniques have not been evaluated in the context of modern LLMs and vision transformers. Given the recent interest in quantization of LLMs, this paper presents an open-source simulator, \textbf{INT-FP-QSim}\footnote{\url{https://github.com/lightmatter-ai/INT-FP-QSim}} for flexible evaluation of LLMs and vision transformers at various numerical formats. INT-FP-QSim combines resources from existing open-source repositories, such as TensorRT \cite{tensorrt}, QPytorch \cite{zhang2019qpytorch} and AIMET \cite{aimet, kuzmin2022fp8} for a combined simulator that enables flexible investigations with a variety of numerical data formats and precisions. With the help of the simulator, we survey the impact of different numerical formats (floating point, integer, mixed floating point and integer) on LLMs and vision transformers, including the application of post-training quantization (PTQ) methods and quantization-aware training (QAT) for improving model performance. In contrast to prior work investigating different PTQ methods specifically for LLMs \cite{yao2023comprehensive}, we cover a range of models (See Figure \ref{fig:model_summary}), and newer PTQ techniques such as Adaptive Block Floating Point (ABFP) \cite{basumallik2022adaptive}, SmoothQuant \cite{xiao2022smoothquant}, GPTQ \cite{frantar2022gptq} and RPTQ \cite{yuan2023rptq}. In all the cases, we investigate 4-bit weights (either integer or floating point), along with 4-bit or 8-bit activations.

We summarize our contributions as follows: a) we present INT-FP-QSim, an open-source flexible simulator for simulating different numerical formats; b) we investigate the impact of using mixed data formats with 4-bit integer weights, and 8-bit floating point activations; and c) we investigate the impact of mixed and low precision on a variety of models and task domains ranging from conventional LLMs like OPT, to transformers for computer vision and text-to-image generation that are often less explored in this context. With INT-FP-QSim and the findings of this work, we hope to recognize insights for future work in this space.

\begin{figure*}[t]
\centering
\includegraphics[width=0.97\textwidth]{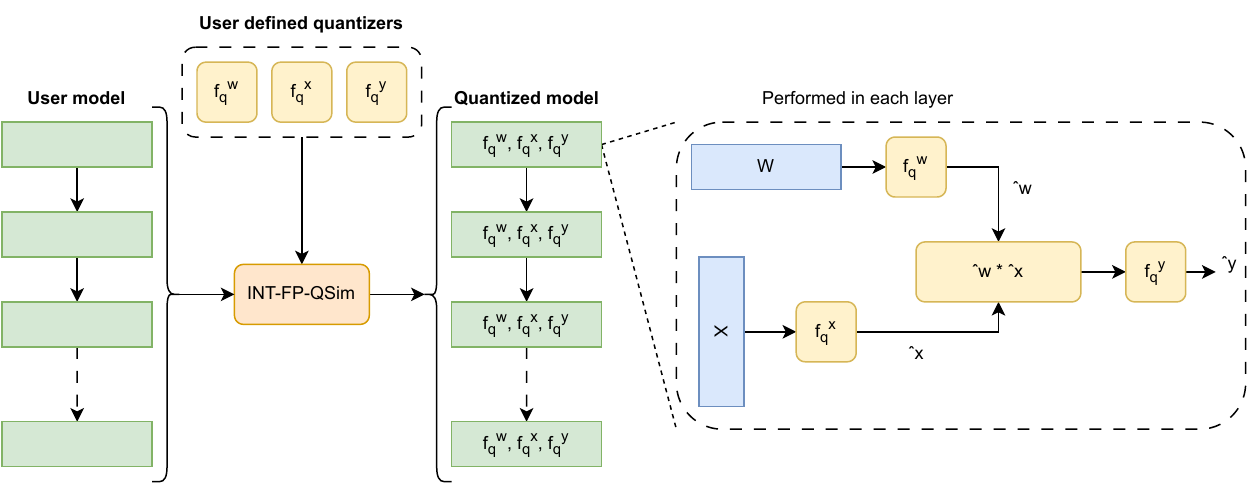}
\caption{High level overview of INT-FP-QSim. The user specifies the input model and the quantizer functions. The matrix multiplication layers (e.g., conv, linear) within the model are replaced with versions of the layers that have the quantizers attached. During forward pass, the quantizers are applied to the inputs, weights and outputs of each layer to simulate quantization.}
\label{fig:workflow}
\end{figure*}

\section{Background and Related Work}
We describe the different numerical formats along with the accuracy recovery methods that we investigate in this work.

\subsection{Numerical Formats}
In this work, we fix weights to 4-bit precision, and explore different precisions for activations. Specifically, with weights kept either in 4-bit integer or 4-bit floating point, we explore activation quantization in the following formats: 1) 4-bit integer; 2) 8-bit integer; 3) 4-bit floating point; and 4) 8-bit floating point \cite{micikevicius2022fp8}. For the 4-bit floating point formats, we explore both 2-bit exponent (with 1-bit mantissa, i.e., E2M1) and 1-bit exponent formats (E1M2). For 8-bit floating point, we use the format proposed in \cite{micikevicius2022fp8}, namely, 4-bit exponent and 3-bit mantissa (E4M3). We perform integer quantization as follows, using nearest rounding:
\begin{gather}
    s = \frac{2^b-1}{\alpha} \\
    x_q = Q(x; s, b) = clip(round(s \cdot x)), -2^b-1, 2^b-1)
\end{gather}
We de-quantize the input as follows:
\begin{gather}
    \hat{x} = DQ(x_q) = \frac{1}{s}x_q
\end{gather}
Performing quantization to lower precision often results in a drop in the model accuracy, requiring the application of special techniques to recover comparably to higher precision levels.

\subsection{Accuracy Recovery With Post-Training Quantization}
We investigate the following methods for accuracy recovery from quantization: a) static quantization with Mean Squared Error (MSE) calibration \cite{wu2020integer}, b) Adaptive Block Floating Point (ABFP) \cite{basumallik2022adaptive}, c) SmoothQuant (SQ) \cite{xiao2022smoothquant}, d) GPTQ \cite{frantar2022gptq}, and e) RPTQ \cite{yuan2023rptq}. We briefly describe each method here, and refer readers to the corresponding papers for more details.

\subsubsection{Static Quantization with MSE Calibration}
Calibration is the process of choosing the scales $\alpha$ (Eqn (1)) for quantizing the model weights and activations \cite{wu2020integer}. The calibrated scales are then used to perform quantization as shown in Eqn (2). In this work, we follow the approach laid out in \cite{wu2020integer} and use per-channel max calibration for weights, with mean-squared error (MSE) calibration for activations. MSE calibration computes the scale (i.e., $\alpha$) that minimizes MSE between the quantized and unquantized outputs of each layer, and has been shown to perform well for minimizing quantization error \cite{sakr2022optimal}.

\subsubsection{Adaptive Block Floating Point}
Recent approaches like VS-Quant \cite{dai2021vs} and Adaptive Block Floating Point (ABFP) \cite{basumallik2022adaptive} have looked at dynamically scaling vectors of length $n$, as a potential solution to maintaining accuracy at low precision. In this work, we focus on ABFP and evaluate its performance for different numerical formats. ABFP computes scale factors as \textit{max} over vectors of length $n$, wherein the scales themselves are left in BF16. Given an $M \times N$ matrix of input activations $X \in \mathbb{R}^{M \times N}$, let $x^i$ represent the $i$-th column of $X$. Then scale factors are computed for vectors of length $n$ in each column:
\begin{align}
    s^i_j = max(x^i_j) \text{ where } x^i_j \in \{x^i_1, x^i_2, \dots, x^i_{\ceil*{\frac{|x^i|}{n}}} \}
\end{align}
The scale $s^i_j$ is used to quantize the $n$-length vector $x^i_j$ of column $x_i$. For weights, Eqn (4) is repeated except the scales are computed over \textit{rows} instead of columns. In contrast to computing the max value over an entire matrix, the computation of max over vectors of length $n$, helps minimize information loss due to quantization. We explore the performance of ABFP for vectors of length $n \in \{64, 128\}$. A second-level quantization for the scale factors could be utilized to achieve further compression \cite{dai2021vs}. However, we do not explore the second-level quantization in this work in order to focus our analysis on the best-case performance of the different models.

\subsubsection{SmoothQuant}
SmoothQuant is a post-training quantization technique shown to work well with 8-bit integer weights and 8-bit integer activations in LLMs \cite{xiao2022smoothquant}. SmoothQuant involves \textit{migrating} the quantization difficulty from activations to weights, based on a smoothing factor. It is motivated by the observation that LLM activations are harder to quantize than weights owing to the presence of a significant amount of outliers. In this work, we explore the application of SmoothQuant (denoted as SQ) for 4-bit weights and 4-bit or 8-bit activations. In our experiments, the migration or smoothing factor is set to 0.5 for all layers within the model based on prior work \cite{xiao2022smoothquant}.

\subsubsection{GPTQ} GPTQ is a post-training quantization approach that utilizes approximate second-order information of the weights in order to compress it to lower precision \cite{frantar2022gptq}. GPTQ achieves improved performance at 4-bit weights and FP16 activations, compared to FP16 baselines. The authors highlight that activations do not pose a significant bottleneck, enabling the 175-billion parameter OPT to be run on a single GPU.

\subsubsection{RPTQ} RPTQ \cite{yuan2023rptq} is a post-training quantization technique that rearranges the channels in activations, and quantizes them in clusters, motivated by the observation that in addition to presence of outliers, varying ranges across channels makes activation quantization difficult. RPTQ demonstrates improved results for 4 and 3-bit quantization of weights and activations.

\subsection{Accuracy Recovery With Training}
Although post-training quantization methods achieve good results, significant improvement is observed with the use of training or fine-tuning methods such as quantization-aware training (QAT). While QAT can be computationally intensive, especially for really large language models, we explore the performance benefits of applying QAT in conjunction with ABFP for models up to OPT 2.7B, in order to assess the degree of performance improvements obtained. During fine-tuning, the forward pass is executed using ABFP, with a Piecewise-linear (PWL) estimator in the backward pass:
\begin{gather}
    \frac{\partial Q(x, s, b)}{\partial x} = \mathbbm{1}_{\{|x| \leq s\}}
\end{gather}

\subsection{Other Related Methods Not Explored In This Work}
In addition to the methods discussed above for low precision execution of LLMs, prior work called LLM-INT8() has looked at running specific channels with a high incidence of outliers in higher precision (FP16), while running inliers in INT8 \cite{dettmers2022llm}. In this work, we do not explore LLM-INT8(), instead focusing on consistently running all the activations and weights in lower precision (4-bit and 8-bit).

\section{Simulator Overview}
In this section, we describe the components of our simulator, INT-FP-QSim, where the high-level workflow is shown in Fig \ref{fig:workflow}.
To begin, the user provides the input model and specifies the input and weight quantizer functions. These functions couple quantization and de-quantization together, to perform \textit{simulated quantization} \cite{wu2020integer} of inputs and weights. The simulator optionally supports quantization of outputs for accommodating alternate hardware configurations. For instance, photonics hardware can involve output quantization operations \cite{demirkiran2021electro}. The quantizer functions perform the following:
\begin{gather}
    \hat{w} = f^w_q(w) = DQ(Q(w; s_w, b_w)) \\
    \hat{x} = f^x_q(x) = DQ(Q(x; s_x, b_x)) \\
    y = \hat{w}*\hat{x} \\
    \hat{y} = f^y_q(y) = DQ(Q(y; s_y, b_y))
\end{gather}
Once the quantizers and input model is specified, the simulator then replaces each layer that performs matrix multiplications (i.e., linear, attention, and convolution) within the model, with versions of the layers that have the quantizers attached to them. The replaced versions of the layers apply the quantizers to the inputs, weights and outputs during the forward pass to simulate quantization. The data-types are maintained in \floatfp, while the data is quantized to lower precision and then de-quantized, as specified in the quantizer configuration by the user. Hence, data is stored in lower precision, while all the non-linear operations and matmuls are performed in higher precision after de-quantization (similar to prior work \cite{frantar2022gptq, micikevicius2022fp8}). This gives us an \textit{upper-bound} on the performance of the models. Our future work will seek to add support for different matmul precisions. The quantizer functions ($f^w_q(w), f^x_q(x), f^y_q(y)$) can be flexibly specified to perform integer or floating point quantization. For instance, the quantizer functions can be specified using TensorRT to perform integer quantization. Alternatively, the quantizer function can be specified using QPytorch in order to perform low precision floating point quantization.


Modified versions of the layers for the models discussed in this work (e.g., \textit{OPTAttention}, \textit{CodegenAttention}) are provided within INT-FP-QSim, to take the quantizer functions that are specified and to perform the simulated quantization operations (Eqns (6)-(9)) during forward pass.

For accuracy recovery, INT-FP-QSim provides support for QAT and static calibration of activations and weights. INT-FP-QSim also includes support for static max calibration of activations, where the max value is pre-calibrated over a subset of data, and statically reused during quantization (in contrast to dynamic max computation). Once the calibration is performed, the quantizer functions use the calibrated scale to perform simulated quantization during the forward pass of the model. Future iterations of our simulator will add ABFP support. PTQ methods such as SmoothQuant, GPTQ and RPTQ are already available as open-source repositories\footnote{SQ: \url{https://github.com/mit-han-lab/smoothquant}; GPTQ: \url{https://github.com/IST-DASLab/gptq}; RPTQ: \url{https://github.com/hahnyuan/RPTQ4LLM}}.

\section{Experiments and Results}
In this section, we investigate the performance of a variety of models on different numerical formats, including the accuracy recovery methods discussed previously. We begin by carrying out a subset of the experiments on four OPT models from OPT 125M - OPT 2.7B on the Wikitext2 dataset, and present more extensive results on additional models for the promising approaches. Note that we are focusing on the best-case outcomes for the models at different precisions, with different techniques. Hence, in all our experiments we do not explore the impact of low-precision output quantization, instead leaving the output activations in FP16. Further, we denote the sequence lengths used for the models in each case. Given the resource constraints for fine-tuning LLMs, we use smaller sequence length of 256 for fine-tuning OPT 1.3B-2.7B.

\textbf{Note on practical implementations of SQ and RPTQ:} When applying SQ and RPTQ from their open source repositories, we note that running SQ for the OPT 350M raised errors owing to slight differences in the model architecture\footnote{See Github issue reported here: \url{https://github.com/huggingface/transformers/issues/17653}}. Hence for the SQ experiments, we report results for OPT 125M, OPT 1.3B and OPT 2.7B. Similarly, the RPTQ implementation does not support OPT 350M and OPT 2.7B, hence RPTQ results for these models are also exempt from the experiments.

\subsection{4-bit weights and 4-bit activations}
In this section, we explore the performance of OPT models for 4-bit weights and 4-bit activations in integer and floating point formats. Specifically, a) we compare static MSE calibration vs. ABFP for 4-bit integer weights and 4-bit integer activations; b) we compare performance of INT4 weights and activations to FP4 weights and activations (in E2M1 and E1M2 formats). Since floating point formats have recently garnered interest \cite{micikevicius2022fp8} we hope to investigate the applicability of 4-bit floating point formats here; c) we look at performance of ABFP, ABFP-QAT (ABFP along with QAT) and ABFP-SQ (ABFP with SmoothQuant) for the different formats.

\subsubsection{Static MSE calibration vs. ABFP}
Table \ref{tab:MSE_vs_abfp} shows the results comparing Static MSE calibration and ABFP for 4-bit integer weights and 4-bit integer activations (W4A4) for OPT 125M and OPT 350M. Given the low bit precisions for the two models, MSE calibration significantly under-performs ABFP, and does not yield a useable perplexity (PPL). This could be correlated to the nature of outliers within these models, wherein the MSE values would have to clip most outliers to be effective, even if the outliers themselves could be critical to model performance. In contrast, ABFP enables the outliers to be represented, while at the same time minimizes information loss due to them by quantizing only vectors of length $n$.

\subsubsection{4-bit integer vs. floating point for weights and activations}
Table \ref{tab:int4_vs_fp4} shows results comparing 4-bit integer weights and activations to 4-bit floating point weights and activations in E2M1 (2-bit Exponent, 1-bit Mantissa) and E1M2 formats. Both formats use ABFP for $n=64$. We see that the 4-bit floating point formats do not offer any significant advantages over 4-bit integers (with the exception of OPT 350M). In particular, the performances of W4A4 and E1M2 seem to be quite similar, while E1M2 under-performs the two. The results also show that ABFP is a viable method that can be applied with either integer or floating point formats.

\begin{figure}[t]
\centering
\includegraphics[width=0.46\textwidth]{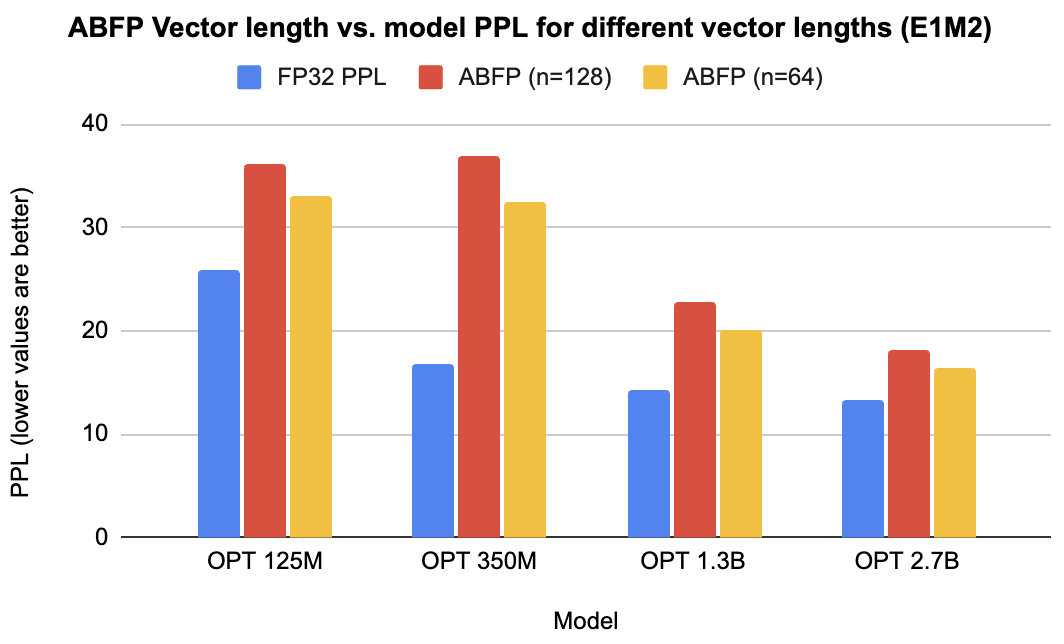}
\caption{Performance of the models with E1M2 weights and activations for different vector lengths $(n=64, 128)$.}
\label{fig:fp4-abfp-ppl}
\end{figure}

In Figure \ref{fig:fp4-abfp-ppl}, we denote the impact of varying the vector lengths $n$ from 64 to 128. We observe that larger vector lengths can impact model performance, but the differences between model PPL from $n=64$ to $n=128$ seem to reduce as model size increases to 1.3B and 2.7B indicating that larger vector lengths could possibly be viable for larger models.

\begin{table}[]
\caption{Perplexity (PPL) values for W4A4 configuration using static MSE calibration and ABFP (n=64). Lower PPL is better.}
\centering
\label{tab:MSE_vs_abfp}
\begin{tabular}{c|c|c|c}
\multicolumn{1}{c|}{Model (Seq len)} & \multicolumn{1}{c|}{FP32} & \multicolumn{1}{c|}{MSE} & \multicolumn{1}{c}{ABFP} \\ \hline
OPT 125M (1024) & 25.94 & 1130 & \textbf{33.14} \\
OPT 350M (1024) & 16.84 & 927 & \textbf{32.28} \\
\end{tabular}
\end{table}

\begin{table}[]
\caption{Comparison of 4-bit integer and 4-bit floating point formats with ABFP (n=64).}
\centering
\label{tab:int4_vs_fp4}
\begin{tabular}{c|c|c|c|c}
\multicolumn{1}{c}{} & \multicolumn{1}{c|}{} & \multicolumn{1}{}{} & \multicolumn{1}{c}{With ABFP} & \multicolumn{1}{}{} \\
\multicolumn{1}{c|}{Model (Seq len)} & \multicolumn{1}{c|}{FP32} & \multicolumn{1}{c|}{W4A4} & \multicolumn{1}{c|}{E2M1} & \multicolumn{1}{c}{E1M2} \\ \hline
OPT 125M (1024) & 25.94 & \textbf{33.14} & 37.59 & \textbf{33.12} \\
OPT 350M (1024) & 16.84 & 32.28 & \textbf{28.55} & 32.41 \\
OPT 1.3B (256) & 14.33 & \textbf{19.97} & 51.82 & 20.06 \\
OPT 2.7B (256) & 13.35 & \textbf{16.39} & 32.77 & 16.41
\end{tabular}
\end{table}

\subsubsection{Accuracy recovery for 4-bit integer weights and activations}
For the training experiments in this section, the backward pass for the function is defined as given in Eqn (5). The results are shown in Table \ref{tab:qat_w_abfp_w4a4}. We see that using ABPF-QAT and ABFP-SQ can both improve performance over ABFP alone. In particular, ABFP-QAT helps achieve PPL values that are close to the baseline model. However, QAT can be highly resource intensive and further investigations into improved fine-tuning methods could be useful. To support the training experiments in our work, we use sequence lengths of 256 for the larger models (OPT 1.3B and OPT 2.7B). For larger models, we see that using ABFP along with SQ can achieve performances that are closer to QAT, indicating that the combination of ABFP and SQ could be a useful technique.

\begin{figure}[t]
\centering
\includegraphics[width=0.46\textwidth]{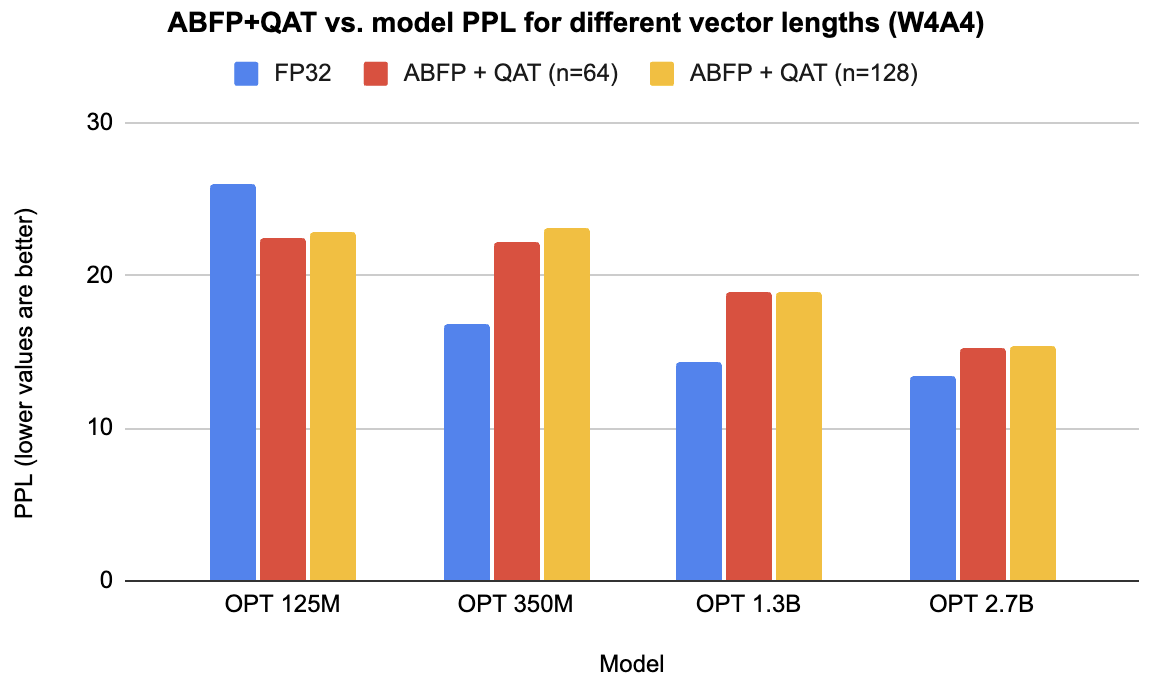}
\caption{Performance of the models with ABFP and QAT for different vector lengths $(n=64, 128)$ in the W4A4 format.}
\label{fig:abfp-qat-ppl}
\end{figure}

We further assess the impact of applying QAT with ABFP for different vector lengths in Figure \ref{fig:abfp-qat-ppl}. In all cases, fine-tuning with QAT helps improve model performance compared to using vanilla ABFP. Additionally, we see that for the different models, applying QAT at larger vector lengths ($n=128$) can achieve performances that are close to smaller vector lengths.

\begin{table}[]
\caption{Accuracy recovery methods on W4A4 representation to augment ABFP for $n=64$.}
\label{tab:qat_w_abfp_w4a4}
\centering
\begin{tabular}{c|c|c|c|c}
\multicolumn{1}{c|}{Model (Seq len)} & \multicolumn{1}{c|}{FP32} & \multicolumn{1}{c|}{ABFP} & \multicolumn{1}{c|}{ABFP-QAT} & \multicolumn{1}{c}{ABFP-SQ} \\ \hline
OPT 125M (1024) & 25.94 & 33.14 & \textbf{22.38} & 32.77 \\
OPT 350M (1024) & 16.84 & 32.28 & \textbf{22.17} & - \\
OPT 1.3B (256) & 14.33 & 19.97 & \textbf{18.97} & 19.57 \\
OPT 2.7B (256) & 13.35 & 16.39 & \textbf{15.22} & 15.78
\end{tabular}
\end{table}

\subsection{4-bit weights and 8-bit activations}
In this section, we explore the performance of OPT models for 4-bit integer weights and 8-bit activations in the integer and floating point formats. Specifically, a) we compare static MSE calibration vs. ABFP for 4-bit integer weights and 8-bit integer activations; b) we compare performance of 8-bit integer activations to FP8 activations (in E4M3 format) with 4-bit integer weights in each case. We also compare the results obtained here to GPTQ baseline; c) we look at performance of ABFP, ABFP-QAT (ABFP along with QAT) and ABFP-SQ (ABFP with SmoothQuant) for the different formats.

\subsubsection{Static MSE calibration vs. ABFP}
Table \ref{tab:MSE_vs_abfp_w4a8} shows the results comparing Static MSE calibration and ABFP for 4-bit integer weights and 8-bit integer activations (W4A8) for OPT 125M - OPT 2.7B. Compared to 4-bit activations, MSE calibration yields a usable PPL with 8-bit activations. Using 8-bit activations enables a wider representation of the outliers, and possibly enables MSE calibration to perform better. However, ABFP outperforms MSE static calibration for all the models, achieving close to the baseline model PPL values.

\begin{table}[]
\caption{W4A8 configuration using static MSE calibration and ABFP (n=64).}
\centering
\label{tab:MSE_vs_abfp_w4a8}
\begin{tabular}{c|c|c|c}
\multicolumn{1}{c|}{Model (Seq len)} & \multicolumn{1}{c|}{FP32} & \multicolumn{1}{c|}{MSE} & \multicolumn{1}{c}{ABFP} \\ \hline
OPT 125M (1024) & 25.94 & 108.98 & \textbf{29.21} \\
OPT 350M (1024) & 16.84 & 29.81 & \textbf{21.67} \\
OPT 1.3B (256) & 14.33 & 36.8 & \textbf{15.54} \\ 
OPT 2.7B (256) & 13.35 & 38.11 & \textbf{14.34}
\end{tabular}
\end{table}

\subsubsection{8-bit integer vs floating point activations}
Table \ref{tab:int4_with_fp8} shows results of using 4-bit integer weights with 8-bit floating point activations (E4M3 format) with ABFP. We also show results of combining ABFP with SmoothQuant and compare it to the GPTQ results (that use FP16 activations and 4-bit weights). We see that using the E4M3 format for activations with ABFP, can help achieve model PPL close to \floatfp{}. Combining ABFP with SmoothQuant can further improve the model PPL, achieving significantly improved results for the OPT models (particularly for OPT 1.3B and OPT 2.7B), compared to the corresponding GPTQ results. However, as shown in Table \ref{tab:int8_vs_fp8}, using E4M3 representation for input activations does not result in any significant improvements over using 8-bit integers.

\begin{table}[]
\caption{Comparison of E4M3 activations and INT4 weights with ABFP ($n=64$) to GPTQ that uses FP16 activations and INT4 weights.}
\centering
\label{tab:int4_with_fp8}
\begin{tabular}{c|c|c|c|c}
\multicolumn{1}{c}{} & \multicolumn{1}{c|}{} & \multicolumn{1}{c}{W4-AE4M3} & \multicolumn{1}{c|}{} & \multicolumn{1}{c}{W4A16} \\\cline{3-5}
\multicolumn{1}{c|}{Model (Seq len)} & \multicolumn{1}{c|}{FP32} & \multicolumn{1}{c|}{ABFP} & \multicolumn{1}{c|}{ABFP-SQ} & \multicolumn{1}{c}{GPTQ} \\ \hline
OPT 125M (1024) & 25.94 & 29.99 & \textbf{29.63} & 30.95 \\
OPT 350M (1024) & 16.84 & 21.99 & - & 19.87 \\
OPT 1.3B (1024) & 11.78 & 13.04 & \textbf{12.54} & 15.13 \\
OPT 2.7B (1024) & 11 & 11.87 & \textbf{11.44} & 14.16
\end{tabular}
\end{table}

\begin{table}[]
\caption{Comparison of E4M3 activations vs. 8-bit integer activations with ABFP ($n=64$) and ABFP with SmoothQuant.}
\centering
\label{tab:int8_vs_fp8}
\begin{tabular}{c|c|c|c|c}
\multicolumn{1}{c|}{} & \multicolumn{1}{l}{W4-AE4M3} & \multicolumn{1}{c|}{} & \multicolumn{1}{c}{W4A8} & \multicolumn{1}{c}{} \\\cline{2-5}
\multicolumn{1}{c|}{Model (Seq len)} & \multicolumn{1}{c|}{ABFP} & \multicolumn{1}{c|}{ABFP-SQ} & \multicolumn{1}{c|}{ABFP} & \multicolumn{1}{c}{ABFP-SQ} \\ \hline
OPT 125M (1024) & 29.99 & 29.63 & 29.21 & \textbf{28.47} \\
OPT 350M (1024) & 21.99 & - & 21.67 & - \\
OPT 1.3B (1024) & 13.04 & 12.54 & 12.73 & \textbf{12.26} \\
OPT 2.7B (1024) & 11.87 & 11.44 & 11.78 & \textbf{11.37}
\end{tabular}
\end{table}

\begin{table}[]
\caption{Accuracy recovery methods on W4A8 representation to augment ABFP for $n=64$.}
\label{tab:qat_w_abfp_w4a8}
\centering
\begin{tabular}{c|c|c|c|c}
\multicolumn{1}{c|}{} & \multicolumn{1}{c}{} & \multicolumn{1}{c}{W4A8} & \multicolumn{1}{c|}{} & \multicolumn{1}{c}{W4A16} \\\cline{2-5}
\multicolumn{1}{c|}{Model (Seq len)} & \multicolumn{1}{c|}{ABFP} & \multicolumn{1}{c|}{ABFP-QAT} & \multicolumn{1}{c|}{ABFP-SQ} & \multicolumn{1}{c}{GPTQ} \\ \hline
OPT 125M (1024) & 29.21 & \textbf{21.22} & 28.47 & 30.95 \\
OPT 350M (1024) & 21.67 & \textbf{18.95} & - & 19.87 \\
OPT 1.3B (256) & 15.54 & \textbf{14.79} & 14.95 & 18.68 \\
OPT 2.7B (256) & 14.34 & \textbf{13.68} & 13.81 & 17.48 
\end{tabular}
\end{table}

\begin{figure}[t]
\centering
\includegraphics[width=0.46\textwidth]{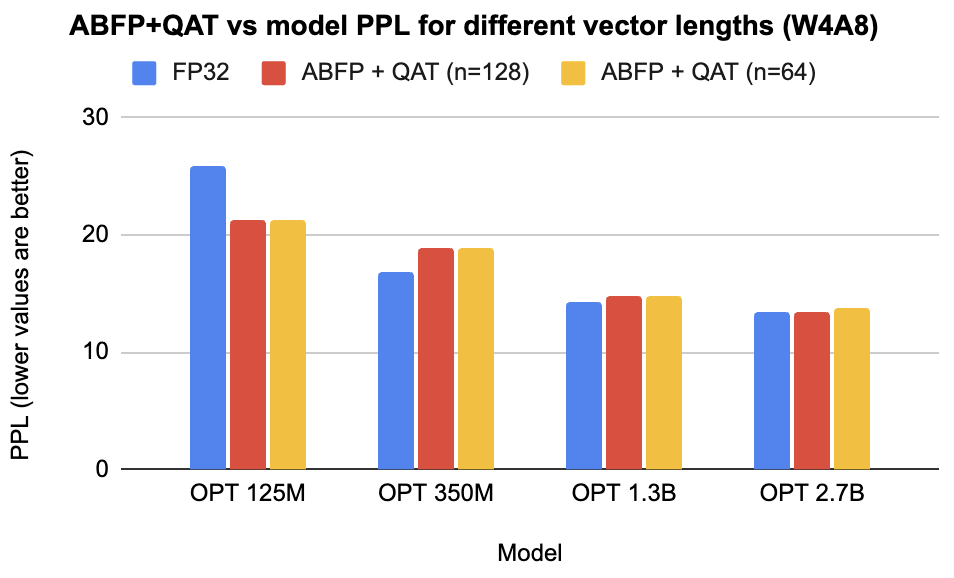}
\caption{Performance of the models with ABFP and QAT for different vector lengths $(n=64, 128)$ in the W4A8 format.}
\label{fig:abfp-qat-w4a8}
\end{figure}

\subsubsection{Accuracy recovery for 4-bit integer weights and 8-bit integer activations}
Results for the experiments in this section are shown in Table \ref{tab:qat_w_abfp_w4a8}. Similar to the case with W4A4 in the previous sections, we see that using ABPF-QAT and ABFP-SQ can both improve performance over ABFP alone. ABFP-QAT helps achieve significantly improved PPL values that are close to the baseline models. In the case of 8-bit activations, ABFP with SmoothQuant also achieves performances that are very close to the baseline models, indicating that QAT does not provide significant advantages here.

We further assess the impact of applying QAT with ABFP for different vector lengths in Figure \ref{fig:abfp-qat-w4a8}. In all cases, fine-tuning with QAT helps improve model performance compared to using vanilla ABFP. Similar to the case with W4A4, we see that for the different models, applying QAT at larger vector lengths ($n=128$) can achieve performances that are very close to smaller vector lengths ($n=64$). Especially in the case of OPT 1.3B and OPT 2.7B the resultant PPL values are very close to the baseline \floatfp{} models.

\subsection{RPTQ vs. ABFP}
In Table \ref{tab:abfp_vs_rptq} we show results comparing RPTQ to ABFP for W4A4 and W4A8 formats. The open-source RPTQ repository does not include support for OPT 2.7B and OPT 350M, hence we report results for OPT 125M and OPT 1.3B. For W4A8, we observe that RPTQ performs slightly better than ABFP for OPT 1.3B, but worse for OPT 125M. However, for W4A4 we see that ABFP performs better than RPTQ. Nevertheless, there is a trade-off regarding storage overhead of the scales for ABFP, which can be further mitigated through a second-order quantization  of the scales themselves \cite{dai2021vs}.

\begin{table}[]
\caption{Comparison of ABFP ($n=64$) and RPTQ for W4A4 and W4A8 formats.}
\centering
\label{tab:abfp_vs_rptq}
\begin{tabular}{c|c|c|c|c|c}
\multicolumn{1}{c}{} & \multicolumn{1}{c|}{} & \multicolumn{1}{r}{W4A4} & \multicolumn{1}{c|}{} & \multicolumn{1}{r}{W4A8} \\\cline{2-6}
\multicolumn{1}{c|}{Model (Seq len)} & \multicolumn{1}{c|}{FP32} & \multicolumn{1}{c|}{RPTQ} & \multicolumn{1}{c|}{ABFP} & \multicolumn{1}{c|}{RPTQ} & \multicolumn{1}{c}{ABFP} \\ \hline
OPT 125M (1024) & 25.94 & 41.48 & \textbf{33.14} & 37.61 & \textbf{29.21} \\
OPT 350M (1024) & 16.84 & - & 32.28 & - & 21.67 \\
OPT 1.3B (1024) & 11.78 & 15.25 & \textbf{14.13} & \textbf{12.54} & 12.73 \\
OPT 2.7B (1024) & 11 & - & 13.68 & - & 11.78
\end{tabular}
\end{table}

\subsection{Results on Additional Models and Domains}
INT-FP-QSim supports simulation of the different models shown in Table \ref{tab:datasets}. The models are shown along with their corresponding tasks, datasets and metrics. In Table \ref{tab:additional_models} we show the results of applying ABFP on 4-bit integer weights with 4-bit and 8-bit integer activations for $n=64$. In the case of W4A8, ABFP enables all the models to achieve good out-of-the-box performance compared to the \floatfp{} baseline. With W4A4, the vision models yield slightly better performances as compared to the large language models indicating that the vision models may be inherently easier to quantize. In particular, the Codegen 6B model drops significantly in performance with 4-bit activations. However, in general, the out-of-the-box performances using ABFP ($n=64$) yields models that perform well, and quite close to \floatfp{} baselines. The performances can be further boosted by using ABFP with other techniques, such as QAT and SmoothQuant. 

\section{Data resources}
In the appendix, we highlight the different baseline model checkpoints used for the experiments (See Table \ref{tab:ckpt} for reference). The source code for INT-FP-QSim is available here: \url{https://github.com/lightmatter-ai/INT-FP-QSim}. We currently do not include the ABFP implementation in the simulator, although it will be added in future iterations.

\begin{table}[]
\caption{Additional models, datasets and tasks investigated in this work along with the corresponding metrics.}
\centering
\label{tab:datasets}
\begin{tabular}{c|c|c|c}
\multicolumn{1}{c|}{Model} & \multicolumn{1}{c|}{Task} & \multicolumn{1}{c|}{Dataset} &  \multicolumn{1}{c}{Metric} \\ \hline
OPT & Language modeling & Wikitext2 & PPL \\
Codegen & Code generation & HumanEval & Pass@1 \\
BERT & Question answering & Squad v1.1 & F1 \\
ViT & Image classification & ImageNet & Accuracy \\
ImageBind & Image classification & ImageNet & Accuracy \\
Maskformer & Image segmentation & ADE20k & mIOU \\
Graphormer & Graph classification & OGBG-Molhiv & Accuracy \\
Stable Diffusion & Text-to-image & Conceptual & CLIP \\
v1.4 & generation & Captions &
\end{tabular}
\end{table}

\begin{table}[t]
\caption{ABFP results on additional models on a series of different tasks for $n=64$ with the W4A4 and W4A8 formats.}
\centering
\label{tab:additional_models}
\begin{tabular}{c|c|c|c}
\multicolumn{1}{c|}{Model (Seq len)} & \multicolumn{1}{c|}{FP32} & \multicolumn{1}{c|}{ABFP; W4A4} & \multicolumn{1}{c}{ABFP; W4A8} \\ \hline
OPT 125M (1024) & 25.94 & 33.14 & 29.21 \\
OPT 350M (1024) & 16.84 & 32.28 & 21.67 \\
OPT 1.3B (1024) & 11.78 & 15.46 & 12.73 \\
OPT 2.7B (1024) & 11 & 13.67 & 11.78 \\ \hline
Codegen 2B & 23.78 & 23.17 & 23.78 \\
Codegen 6B & 26.83 & 21.95 & 23.78 \\ \hline
BERT-base & 86.81 & 83.57 & 86.12 \\
BERT-large & 93.48 & 92.41 & 93.14 \\ \hline
ViT-large-16 & 79.7 & 79 & 79.7 \\
ViT-large-32 & 77 & 75 & 76.7 \\ \hline
ImageBind & 72.5 & 70.4 & 71.6 \\ \hline
Maskformer-base & 52.85 & 51.29 & 52.71 \\
Maskformer-large & 54.37 & 51.75 & 53.83 \\ \hline
Graphormer & 98.27 & 98.18 & 98.27 \\ \hline
Stable Diffusion v1.4 & 30.51 & 28.51 & 30.17
\end{tabular}
\end{table}

\section{Discussion}
In this work, we presented INT-FP-QSim, an open-source simulator for flexible evaluation of large language models and vision transformers. INT-FP-QSim leverages existing open-source repositories like TensorRT, QPytorch and AIMET to provide a single destination for conducting experiments with models for different numerical formats. With the simulator, we demonstrate results for a range of models, tasks and precisions, and numerical formats.

With the existing implementation, we currently only analyze the models from an accuracy perspective. Actual hardware implementations of the different numerical formats can involve additional challenges in terms of throughput and latency, that is not covered in this work. Further, matmuls are done in \floatfp{} precision to understand the best-case performance of the different models. However, INT-FP-QSim currently does not support matmuls in other formats. Lastly, INT-FP-QSim currently does not support specification of different quantizers for different layers in the model. Nevertheless, we hope that access to INT-FP-QSim will enable researchers to easily and flexibly simulate models at various precision combinations of weights, activations (including output activations if desired), to encourage further research and development in quantization of LLMs and vision transformers.

\section{APPENDIX}
In Table \ref{tab:ckpt} we provide details of the different checkpoints used in our experiments. Most of the checkpoints are official HuggingFace checkpoints, while ViT is obtained from TorchVision. Instead of using the original Facebook checkpoints for OPT that are not fine-tuned on Wikitext2, we used available checkpoints of the models that were specifically fine-tuned on Wikitext2 or OGBG-Molhiv for Graphormer.
\begin{table}[h!]
\caption{Checkpoint links for the different models used in the experiments. HF denotes HuggingFace checkpoints.}
\centering
\label{tab:ckpt}
\begin{tabular}{c|c}
\multicolumn{1}{c|}{Model} & \multicolumn{1}{c}{Checkpoint Details} \\ \hline
OPT 125M & HF: Aalaa/opt-125m-finetuned-wikitext2 \\
OPT 350M & HF: lnair/opt-350m-wikitext2 \\
OPT 1.3B & HF: lnair/opt-1.3b-wikitext2 \\
OPT 2.7B & HF: lnair/opt-2.7b-wikitext2 \\
Codegen 2B & HF: Salesforce/codegen-2B-mono \\
Codegen 6B & HF: Salesforce/codegen-6B-mono \\
BERT-base & HF: deepset/bert-base-cased-squad2 \\
BERT-large & HF: bert-large-cased-whole-word-masking-finetuned-squad \\
ViT-16, 32 & Torchvision: vit\_l\_16, vit\_l\_32 \\
ImageBind & Github: facebookresearch/ImageBind \\
Maskformer-base & HF: facebook/maskformer-swin-base-ade \\
Maskformer-large & HF: facebook/maskformer-swin-large-ade \\
Graphormer & HF: lnair/graphormer-ogbg-molhiv \\ 
Stable Diffusion & HF: CompVis/stable-diffusion-v1-4
\end{tabular}
\end{table}


\bibliography{main}
\bibliographystyle{unsrt}

\end{document}